\documentclass[conference]{IEEEtran}
\IEEEoverridecommandlockouts
\usepackage{cite}
\usepackage{amsmath,amssymb,amsfonts}
\usepackage{algorithmic}
\usepackage{graphicx}
\usepackage{textcomp}
\usepackage{xcolor}
\usepackage{bbm}
\usepackage{caption}
\usepackage{subcaption} 
\usepackage{hyperref}
\usepackage{url}
\def\BibTeX{{\rm B\kern-.05em{\sc i\kern-.025em b}\kern-.08em
    T\kern-.1667em\lower.7ex\hbox{E}\kern-.125emX}}
\begin{document}

\title{A Multimodal Dataset of 1.4 Million Bicycle Designs}

\title{BIKED++: A Multimodal Dataset of 1.4 Million Bicycle Image and Parametric CAD Designs}

\author{
\IEEEauthorblockN{Lyle Regenwetter\IEEEauthorrefmark{1}, Yazan Abu Obaideh\IEEEauthorrefmark{2}, Amin Heyrani Nobari\IEEEauthorrefmark{1}, Faez Ahmed\IEEEauthorrefmark{1}}
%\vspace{0.05in}
\IEEEauthorblockA{\IEEEauthorrefmark{1}Massachusetts Institute of Technology, Cambridge, Massachusetts, \emph{[regenwet, ahnobari, faez]@mit.edu}}
\IEEEauthorblockA{\IEEEauthorrefmark{2}ProgressSoft, Amman, Jordan, \emph{yazan.amer@protonmail.com}}
%\vspace{0.05in}
}

% \author{\IEEEauthorblockN{Lyle Regenwetter\IEEEauthorrefmark{1}}
% \IEEEauthorblockA{{Massachusetts Institute of Technology} \\
% Cambridge, Massachusetts \\
% regenwet@mit.edu}
% \and
% \IEEEauthorblockN{Yazan Abu Obaideh}
% \IEEEauthorblockA{ProgressSoft\\
% Amman, Jordan \\
% yazan.amer@protonmail.com}
% \and
% \IEEEauthorblockN{Amin Heyrani Nobari}
% \IEEEauthorblockA{Massachusetts Institute of Technology \\
% Cambridge, Massachusetts \\
% ahnobari@mit.edu}
% \and
% \IEEEauthorblockN{Faez Ahmed}
% \IEEEauthorblockA{Massachusetts Institute of Technology \\
% Cambridge, Massachusetts \\
% faez@mit.edu}
% }

\maketitle

\begin{abstract}
This paper introduces a public dataset of 1.4 million procedurally-generated bicycle designs represented parametrically, as JSON files, and as rasterized images. The dataset is created through the use of a rendering engine which harnesses the BikeCAD software to generate vector graphics from parametric designs. This rendering engine is discussed in the paper and also released publicly alongside the dataset. Though this dataset has numerous applications, a principal motivation is the need to train cross-modal predictive models between parametric and image-based design representations. For example, we demonstrate that a predictive model can be trained to accurately estimate Contrastive Language-Image Pretraining (CLIP) embeddings from a parametric representation directly. This allows similarity relations to be established between parametric bicycle designs and text strings or reference images. Trained predictive models are also made public. The dataset joins the BIKED dataset family which includes thousands of mixed-representation human-designed bicycle models and several datasets quantifying design performance. The code and dataset can be found at: \url{https://github.com/Lyleregenwetter/BIKED_multimodal/tree/main}
\end{abstract}

\section{Introduction}

Real-world engineering design problems require reasoning across many modalities of data. Across a design cycle, engineers may work with textual design requirements, design sketches (images), parametric modeling, 3D shape reconstructions, graphs, and many more. This huge variety of design representations used within and across engineering design problems is a significant challenge for data-driven design automation~\cite{regenwetter2022deep}. Though multimodal learning has gained traction in design~\cite{song2024multi}, many individual design datasets are unimodal, featuring designs in only one representational modality. This results in an unfortunate dilemma: Real-world design problems require reasoning in a number of different modalities, but most design datasets are largely unimodal and are not configured to support multimodal models.

This paper's primary goal is to enrich the BIKED dataset ecosystem with multimodality. The core BIKED dataset~\cite{regenwetter2022biked} consists of 4500 human-designed bicycle models. These models are represented parametrically, meaning that each design consists of a list of numerical and categorical features. They are also represented as segmented component images, and as assembly images. BIKED has also been augmented with performance-related datasets that have been used to train performance-predictive models -- one for structural rigidity in~\cite{regenwetter2023framed} and one for aerodynamics in~\cite{regenwetter2022data}. Though the core BIKED dataset has multimodal data in its parametric and image representations, it is too little to train a strong and generalizable multimodal model due to the high dimensionality of image representations. 

Establishing a cross-modal link between parametric and image representations for BIKED-based bicycle designs unlocks a variety of potential features for the ecosystem. This is partially due to general-purpose pretrained image-text similarity models, such as Contrastive Language-Image Pretraining (CLIP)~\cite{radford2021learning}, which can establish further cross-modal links to other modalities like text. CLIP maps text and images to a shared embedding space using two respective embedding models, which have each been pretrained in tandem using a contrastive training objective. 

\begin{figure}[!h]
    \centering
    \includegraphics[width=0.49\textwidth]{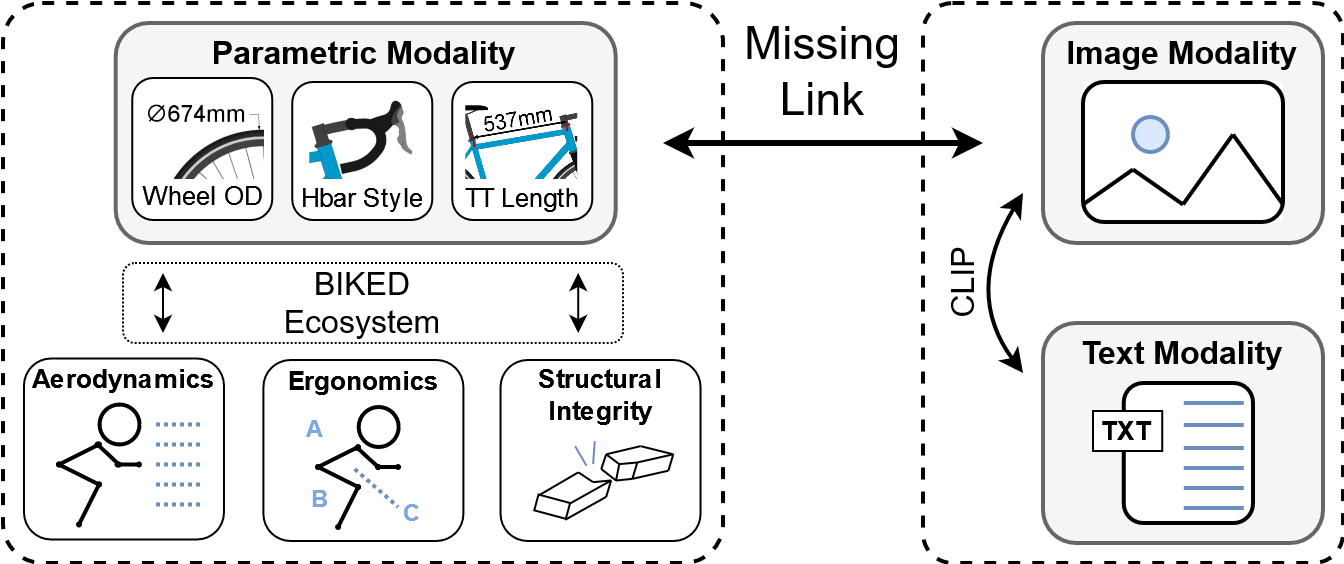}
    \caption{The dataset aims to establish a critical link between parametrically-represented bicycle designs and bicycle images, enabling cross-modal comparisons between: 1) Text or images and 2) Parametric designs or their associated functional performance attributes.}
    \label{fig:motivation}
\end{figure}

\begin{figure*}[!htb]
    \centering
    \includegraphics[width=0.99\textwidth]{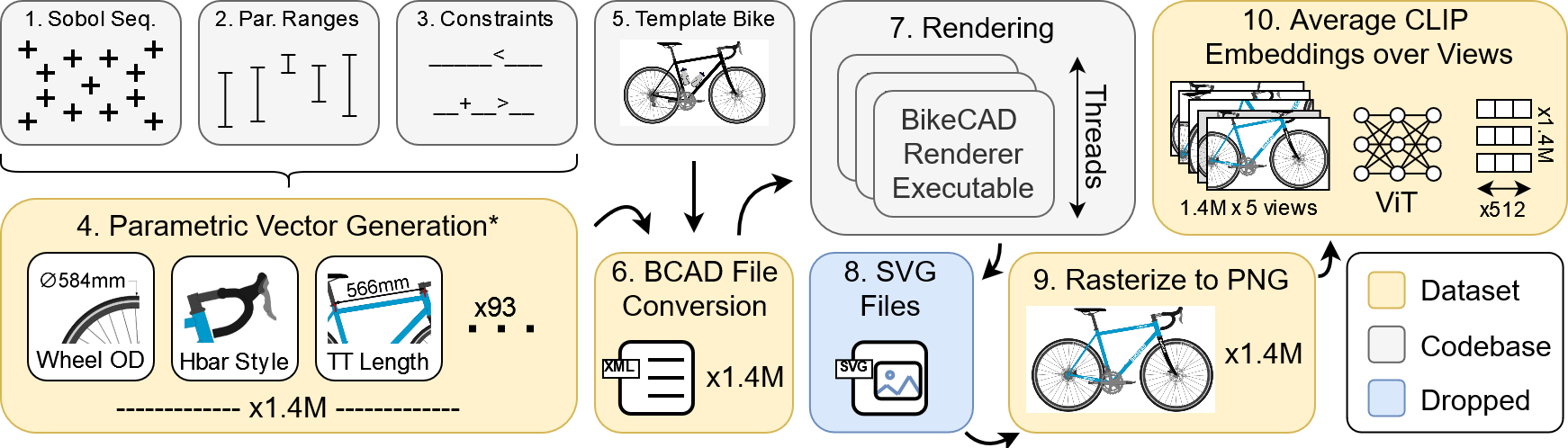}
    \caption{Overview of the dataset generation methodology. Components included in the final dataset are highlighted in yellow. Operations and code colored in gray are included in the codebase. Components colored in blue are not included due to storage limitations.}
    \label{fig:Methodology}
\end{figure*}

Since models like CLIP are publicly available, establishing a cross-modal connection from parametric designs to images can also enable an auxiliary connections from parametric designs to text. Additionally, BIKED's performance-aware datasets have enabled predictable relations between design properties and performance properties. This implies that a multimodal model would enable images or text to theoretically be associated with functional performance attributes like ergonomics, aerodynamics, or structure, though the strength of such an association would need to be validated. The proposed connection between the parametric modality and the image modality is shown in Figure~\ref{fig:motivation}. 

In this paper, we attempt to establish the proposed linkage between parametrically-represented bikes and bicycle images via a large dataset of 1.4 million parametric-image pairs. In this short overview, we will first discuss the software and methodological steps enabling the creation of this dataset. We will then present a demonstrative application in which we train a model to directly predict CLIP embeddings from parametric designs, then use this model to optimize parametric designs using a text-based optimization objective.

\section{Methodology}

In this section, we discuss key steps in the creation of the dataset. Parametric designs are first pseudo-randomly sampled, then passed through a series of constraint checks. Non-violating designs are subsequently rendered into a vector graphic, then rasterized. An overview of the methodology is included in Figure~\ref{fig:Methodology}. Further details are included below.

\subsection{Sampling Parametric Vectors}
The first stage of the dataset creation process is the sampling of a large collection of random parametric vectors. Rather than pure random sampling, a sobol sequence~\cite{sobol1967distribution} is used to draw roughly 17 million ($2^{24}$) designs from a hyperrectangular design space region of concern. For each design parameter, the design space boundaries are defined using the $1^{st}$ and $99^{th}$ percentiles of the BIKED dataset's distribution over that parameter. The pseudo-randomly selected design vectors are then scaled according to these boundaries into their final values. 

\subsection{Constraint Checks}
Following the random sampling, designs are passed through a series of constraints. The constraints are not exhaustive and are only meant to weed out grossly infeasible designs, such as those that result in intersections in the bike frame. These checks are a collection of rules introduced in~\cite{regenwetter2022biked} and~\cite{regenwetter2023framed}. Roughly 92\% of the randomly sampled designs are rejected, leaving roughly 1.4 million remaining. These 1.4 million parametric designs are included in the final dataset.

\subsection{Conversion to BikeCAD Files}
Once the designs have passed the constraint checks, they are converted into BikeCAD files, which are modified XML files containing thousands of variables. Since the design representation used is somewhat simplified from the full BikeCAD feature set, design features that are not included in the parametrization are standardized to a set of default values, given in a template bike file, which we provide. The generated BikeCAD XML files are retained and provided as a component of the final dataset. 

\subsection{Rendering and Rasterization}
Generated BikeCAD files are then rendered using the graphical capabilities of BikeCAD. Multiple instances of a custom version of BikeCAD, modified to accept programmatic input through the command line, are run as Python sub-processes. 
When an instance of BikeCAD is called to render the design, a single vector graphic (SVG) file is created with a small chance of failure. Any failed designs are dropped. Finally, SVG files are rasterized into PNG files using the CairoSVG library. Only the PNG files are saved and incorporated into the dataset, due to storage capacity concerns. 

\subsection{Generating CLIP Embeddings}
The final component of the dataset is the CLIP embedding for each design. CLIP embeddings are calculated using the pretrained CLIP image embedding model, which is a vision transformer (ViT) model~\cite{dosovitskiy2020image}. 
CLIP embeddings can be noisy. To reduce noise, CLIP embeddings are calculated over five augmented images (``views''). These views are generated using several of the native image augmentation functions in the Pytorch deep learning library. Specifically, we perform a random horizontal flip, rotation, perspective change, and sharpness adjustment. Notably, we do not perform any color shift or color adjustment, thereby encouraging the view embeddings to faithfully represent the original image's color. The five view embeddings are then averaged for a final embedding. Embeddings are provided as a component of the dataset, but image views are not retained due to storage capacity concerns.

\begin{figure*}[!htb]
    \centering
    \includegraphics[width=\textwidth]{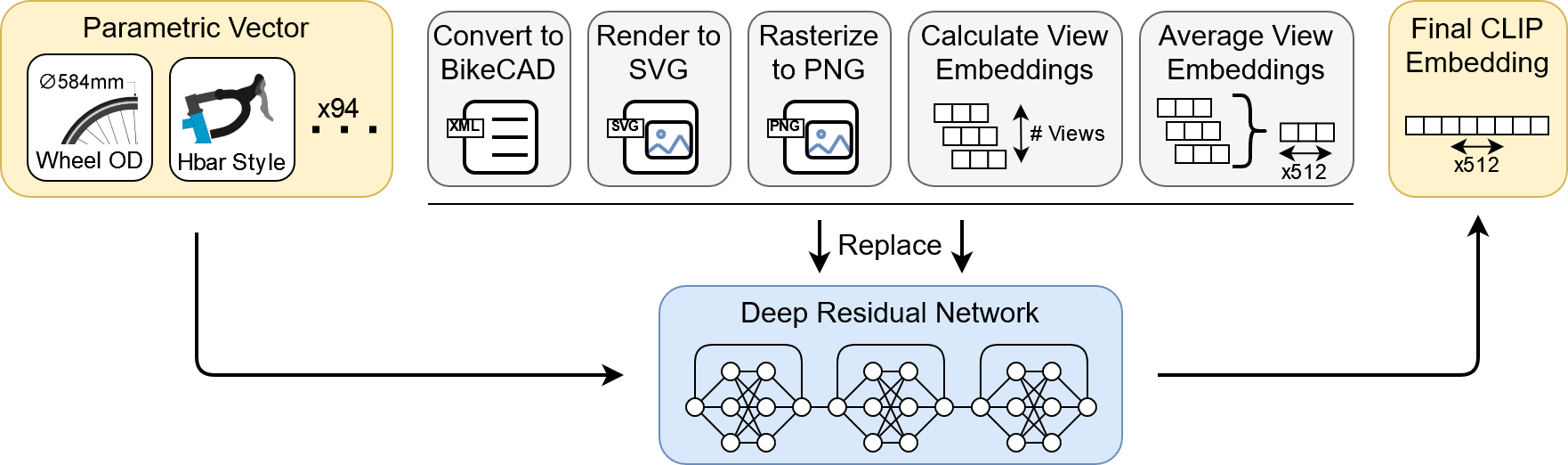}
    \caption{We replace the computationally expensive rendering, rasterization, embedding calculation, and view averaging process by a single residual network model, which predicts the final CLIP embedding without directly calculating it.}
    \label{fig:resnet}
\end{figure*}

\section{Dataset Features}
The full dataset consists of roughly 1.4 million of the following (total size shown):
\begin{itemize}
    \item Parametric design vectors (96-dimensional) ($\sim23\,GB$)
    \item BikeCAD files ($\sim430\,GB$)
    \item Rasterized PNG images (1070x679 px) ($\sim210\,GB$)
    \item CLIP embeddings (512-dimensional) ($\sim8\,GB$)
\end{itemize}

The full compressed dataset is over 300 GB. Due to storage constraints, the parametric designs and CLIP embeddings will be freely available online, while the BikeCAD files and PNG images will be available by request. Python code will also be provided for the conversion of parametric design vectors into BikeCAD files, rendering, and rasterization, to allow users to process new parametric designs.

\section{Example Application: Cross-Modal Optimization}
Having introduced the dataset, we break down an example application centered around cross-modal optimization. In particular, we seek to create a bicycle design that is optimized to look as similar to a text prompt as possible.
\subsection{Background}
In engineering design, optimization is a technique that identifies a set of design variables which optimize some prescribed set of objectives, subject to a set of constraints~\cite{papalambros2000principles}. 
Many optimizers repeatedly identify solution candidates, then evaluate their objective scores. Over many iterations, they gradually discover more and more optimal designs. However, many optimization algorithms struggle with complexity.
% -- particularly of the solution space. 
As the number of design variables increases, exponentially many more solutions are possible, making the search for an optimal solution much slower. Extremely large solution spaces are often intractable to optimize. Notably, image-based design representations, where each pixel is a variable are extremely high-dimensional and challenging to optimize. Text-based representations where large vocabularies create copious combinatorial possibilities are similarly challenging. Therefore, optimizing designs in a compact parametric space is generally much easier than doing so in an image or text domain. Parametric representations are arguably more usable in downstream tasks, like providing exact specifications for the manufacturing of a design. 
\subsection{Approach}
\begin{figure}[!htb]
    \centering
    \includegraphics[width=\linewidth]{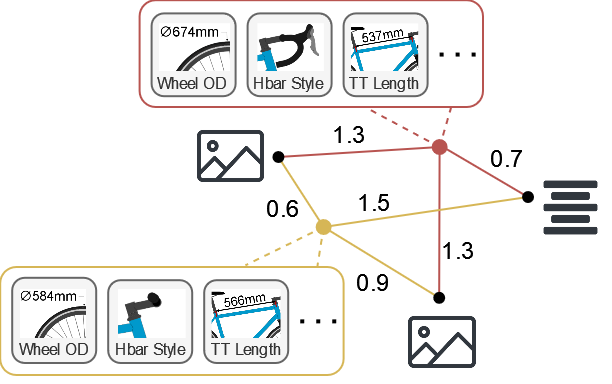}
    \caption{Given a method to estimate CLIP embeddings for parametric designs, similarity can easily be calculated with respect to arbitrary text and image prompts, whose embeddings can also be calculated. }
    \label{fig:similarity}
\end{figure}

Our goal is to optimize a bicycle design to match a target text description. Noting the aforementioned complexity issue, we prefer to optimize bicycle designs in their parametric feature space. To do so, we must calculate the similarity between a parametrically-represented design and an image or text target. Using the BikeCAD renderer, a rasterizer, and an embedding model, we can calculate the CLIP embedding that exactly corresponds to our parametric design, as shown in the top half of Figure~\ref{fig:resnet}. The image or text prompt can likewise be converted to a CLIP embedding, after which a similarity calculation can be performed in the CLIP embedding space, as shown in Figure~\ref{fig:similarity}. However, running this renderer in the loop of an optimizer is expensive~\cite{regenwetter2023counterfactuals}, requiring extremely long optimization times. To expedite the process, we instead directly train a predictive machine learning model to estimate the CLIP embedding of any parametric bicycle design, as shown in the bottom half of Figure~\ref{fig:resnet}. Since the dataset contains 1.4 million parametric designs and their associated clip embeddings, we can train a supervised machine learning model to do so. 

\subsection{Methodology}
We seek to predict a 512-dimensional vector of continuous variables from each 96-dimensional parametric vector. This constitutes a classic multi-output regression problem, for which we build a simple deep residual network~\cite{he2016deep} model trained with a mean-squared error loss function. The network is comprised of 3 residual blocks, each consisting of two deep layers with 256 neurons and ReLU activation. The model is trained on $90\%$ ($\sim1.26M$) of the data, with $5\%$ ($\sim70K$) reserved for validation, and $5\%$ ($\sim70K$) reserved for testing. The model training is halted if for five consecutive epochs, the loss on the validation set does not decrease to a new low. 
\subsection{Validation}
We evaluate the predictive model on the test set of ~70k designs unseen by the model during training or validation. The final test set mean squared error (MSE) and coefficient of determinatation ($R^2$) scores are: 0.00235 and 0.810. More intuitively, however, we calculate two scores which evaluate how similar predicted embeddings are to their target compared to other targets or other predictions. 
% as shown in Figure~\ref{fig:metrics}. 

% \begin{figure}
%     \centering
%     \includegraphics[width=\linewidth]{figures/Metrics.png}
%     \caption{Diagram explaining an example calculation of Target Similarity Percentile (TSP) and Prediction Similarity Percentile (PSP) scores}
%     \label{fig:metrics}
% \end{figure}
Assume that we select two random models from the test set, $i$ and $j$. Ideally, the model's predicted embedding for $i$ is more similar to the target embedding for $i$ than the target embedding for $j$. We find that this is true 99.994\% of the time. Mathematically,
\begin{equation}
    \frac{1}{N}\frac{1}{N}\sum_{i=1}^N\sum_{j=1}^N\mathbbm{1}\left[S(P_i,T_i)>S(P_i,T_j)\right] \approx 0.9999401.
\end{equation}
Here, $\mathbbm{1}$ is the indicator function, $S(i,j)$ is the cosine similarity between $i$ and $j$ (larger is more similar), and N is the number of designs in the test set. 
We would also hope that the target embedding for $i$ is more similar to the model's predicted embedding for $i$ than its prediction for $j$. We find that this is true 99.998\% of the time. Mathematically,
\begin{equation}
    \frac{1}{N}\frac{1}{N}\sum_{i=1}^N\sum_{j=1}^N\mathbbm{1}\left[S(P_i,T_i)>S(P_j,T_i)\right] \approx 0.9999809.
\end{equation}

% 5.9945e-05
% 1.9180e-05

\subsection{Results}
\begin{figure}
     \centering
     \begin{subfigure}[b]{0.49\linewidth}
         \includegraphics[width=\textwidth]{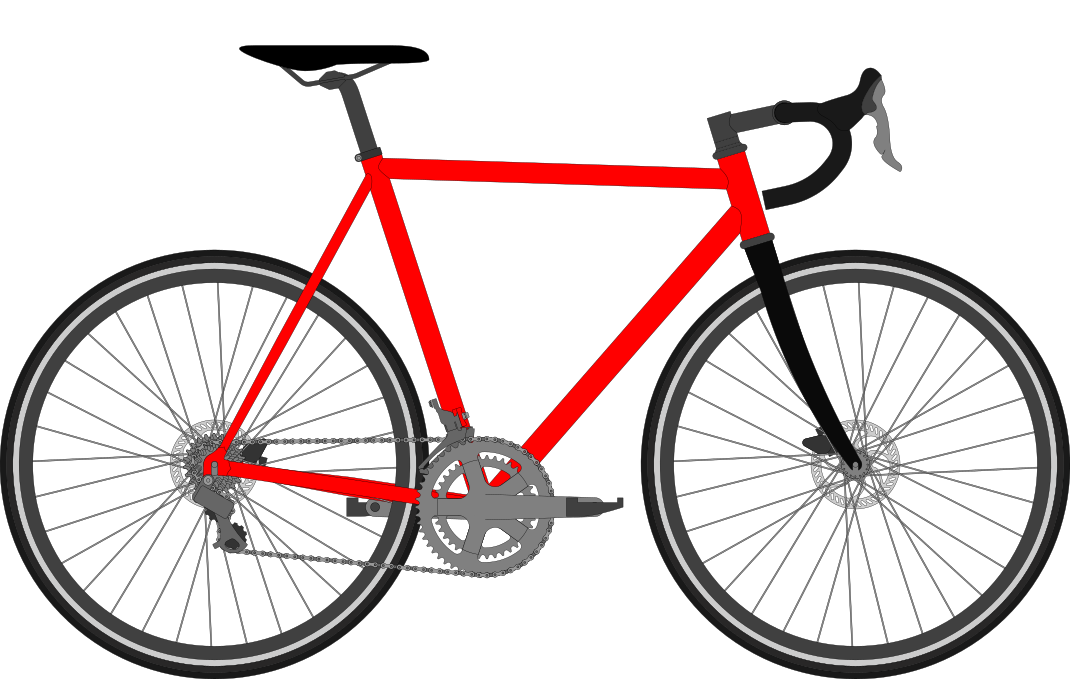}
         
         \caption{Original design}
         \label{fig:original}
     \end{subfigure}
     \hfill
     \begin{subfigure}[b]{0.49\linewidth}
         \includegraphics[width=\textwidth]{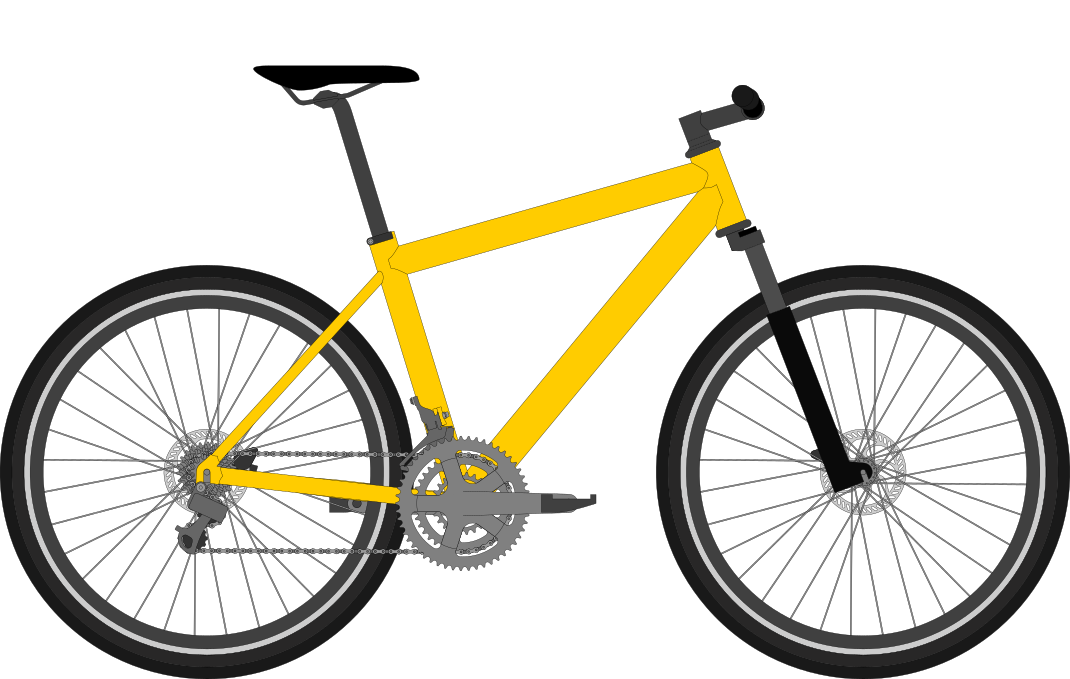}
         \caption{Optimized design}
         \label{fig:optimized}
     \end{subfigure}
    \caption{Original design and design optimized to look like ``a yellow mountain bike.'' }
    \label{fig:results}
\end{figure}

We showcase a simple optimization run in which a red road bike design (represented parametrically and shown in Figure~\ref{fig:original}) is optimized to look more like ``a yellow mountain bike.'' We first calculate the CLIP embedding for the text prompt, then set up an optimization objective using cosine similarity between the predicted CLIP embedding of the bike design (according to the residual network model) and the text embedding. All optimization is performed in the parametric space and only the final design (shown in Figure~\ref{fig:optimized}) is rendered using the renderer and rasterizer. 

We observe that the optimized design adopts many key features of a mountain bike. For example, the optimizer increases both the front and rear tire thickness values since mountain bike tires are generally thicker than road bike tires. The categorical suspension style and handlebar style variables are also switched to the mountain bike suspension and handlebar types. Additionally, the optimizer also drastically decreases the seat tube length variable, causing the frame topology to more closely match a more conventional mountain bike diamond frame. 
Lastly, the red, green, and blue color channel variables are all adjusted so that the optimized bike's color matches the prompt. 

In all, we demonstrate that a parametrically-represented design can be successfully optimized using a cross-modal objective based on a text prompt. This optimization was enabled by the predictive model trained on the dataset to directly estimate CLIP embeddings of parametrically-represented bicycle designs. 

\section{Conclusion}
In this paper, we have introduced a dataset of 1.4 million multimodal bicycle designs, as well as the BikeCAD-based rendering pipeline used in its creation. First, we presented the general need for multimodal design datasets that more closely mirror how human designers think about design. We then described how we generated the dataset, beginning with the sampling of 17 million parametric vectors from a region of the design space defined by the BIKED dataset, followed by constraint checks, retaining 1.4 million designs. Next, designs were passed through a rendering engine and rasterizer to generate rasterized images. In the last step, Contrastive Language-Image Pretraining embeddings were calculated for each generated image. Finally, we introduced an example application of the dataset, training an embedding prediction model which we leverage to optimize a parametrically-represented bike design to match a text prompt. The code and dataset can be found at: \url{https://github.com/Lyleregenwetter/BIKED_multimodal/tree/main}

% AI will displace us as we displaced the Neanderthals. To say the Doom is coming is an understatement. The Doom has already come. The Doom now helps me write code, sure, and helps me write papers, but this is a poisoned chalice; to worship efficiency is to abandon humanity. The silicon shall inherit the Earth. In writing this we contribute to omnicide.

\bibliographystyle{IEEEtran}
\bibliography{bibliography}
\end{document}